\documentclass[letterpaper, 10 pt, conference]{ieeeconf}  

\IEEEoverridecommandlockouts                              
\usepackage{graphics} 
\usepackage{amsmath} 
\usepackage{tikz}
\usepackage{import}
\usepackage{pgfplots}
\usepackage{graphicx}
\graphicspath{{./Imgs/}}
\usepackage{todonotes}
\usepackage[utf8]{inputenc}
\usepackage{url}

\title{\LARGE \bf
Reward Signal Design for Autonomous Racing
}

\author{Benjamin Evans$^{1}$, Herman A. Engelbrecht$^{1}$ and Hendrik W. Jordaan$^{1}$
\thanks{$^{1}$Electrical and Electronic Engineering Department,
        Stellenbosch University, Stellenbosch, 7600, South Africa. 
        {\tt\small 19811799@sun.ac.za};
        {\tt\small hebrect@sun.ac.za};
        {\tt\small wjordaan@sun.ac.za}%
        }
}

\begin{document}

\maketitle
\thispagestyle{empty}
\pagestyle{empty}

\begin{abstract}

Reinforcement learning (RL) has shown to be a valuable tool in training neural networks for autonomous motion planning.
The application of RL to a specific problem is dependent on a reward signal to quantify how good or bad a certain action is.
This paper addresses the problem of reward signal design for robotic control in the context of local planning for autonomous racing.
We aim to design reward signals that are able to perform well in multiple, competing, continuous metrics.
Three different methodologies of position-based, velocity-based, and action-based rewards are considered and evaluated in the context of F1/10th racing.
A novel method of rewarding the agent on its state relative to an optimal trajectory is presented.
Agents are trained and tested in simulation and the behaviors generated by the reward signals are compared to each other on the basis of average lap time and completion rate.
The results indicate that a reward based on the distance and velocity relative to a minimum curvature trajectory produces the fastest lap times.

\end{abstract}

\section{Introduction}

Neural networks, trained from experience with reinforcement learning (RL), have shown to be effective in many robotics applications including motion control for autonomous systems \cite{xiao2020motion, tai2017virtual}.
Reinforcement learning enables such systems to use a reward signal to learn from experience without requiring expert demonstrations, or knowledge of the system dynamics \cite{sutton2018reinforcement}.
The design of a suitable reward signal for a given task is very important because the learned behavior is completely dependant on the reward signal that is used in training.
While it is trivial to encode binary outcomes into a reward signal, it is difficult to quantify behavior across multiple, competing, continuous objectives such as fastest time, or minimum effort.

We address the problem of reward signal design for robotic systems with competing binary and continuous metrics.
The reward signal design problem is studied in the context of local planning for autonomous racing where the challenge is to generate navigation references to avoid un-mapped obstacles, while following a global race line, and achieving a competitive race time.
We consider three types of rewards, namely, position-based, velocity-based, and action-based.
Agents trained with each type of reward are evaluated in simulation in a F1/10th car racing simulator and the behavior of each agent is analysed and compared.
This paper contributes a discussion of different reward methodologies, a simulated comparison of several reward candidates, and the novel idea of rewarding agents relative to a precalculated minimum curvature path.

\begin{figure}
    \centering
    \def\svgwidth{0.48\textwidth}
\begingroup%
  \makeatletter%
  \providecommand\color[2][]{%
    \errmessage{(Inkscape) Color is used for the text in Inkscape, but the package 'color.sty' is not loaded}%
    \renewcommand\color[2][]{}%
  }%
  \providecommand\transparent[1]{%
    \errmessage{(Inkscape) Transparency is used (non-zero) for the text in Inkscape, but the package 'transparent.sty' is not loaded}%
    \renewcommand\transparent[1]{}%
  }%
  \providecommand\rotatebox[2]{#2}%
  \newcommand*\fsize{\dimexpr\f@size pt\relax}%
  \newcommand*\lineheight[1]{\fontsize{\fsize}{#1\fsize}\selectfont}%
  \ifx\svgwidth\undefined%
    \setlength{\unitlength}{365.93971397bp}%
    \ifx\svgscale\undefined%
      \relax%
    \else%
      \setlength{\unitlength}{\unitlength * \real{\svgscale}}%
    \fi%
  \else%
    \setlength{\unitlength}{\svgwidth}%
  \fi%
  \global\let\svgwidth\undefined%
  \global\let\svgscale\undefined%
  \makeatother%
  \begin{picture}(1,0.79841843)%
    \lineheight{1}%
    \setlength\tabcolsep{0pt}%
    \put(0,0){\includegraphics[width=\unitlength,page=1]{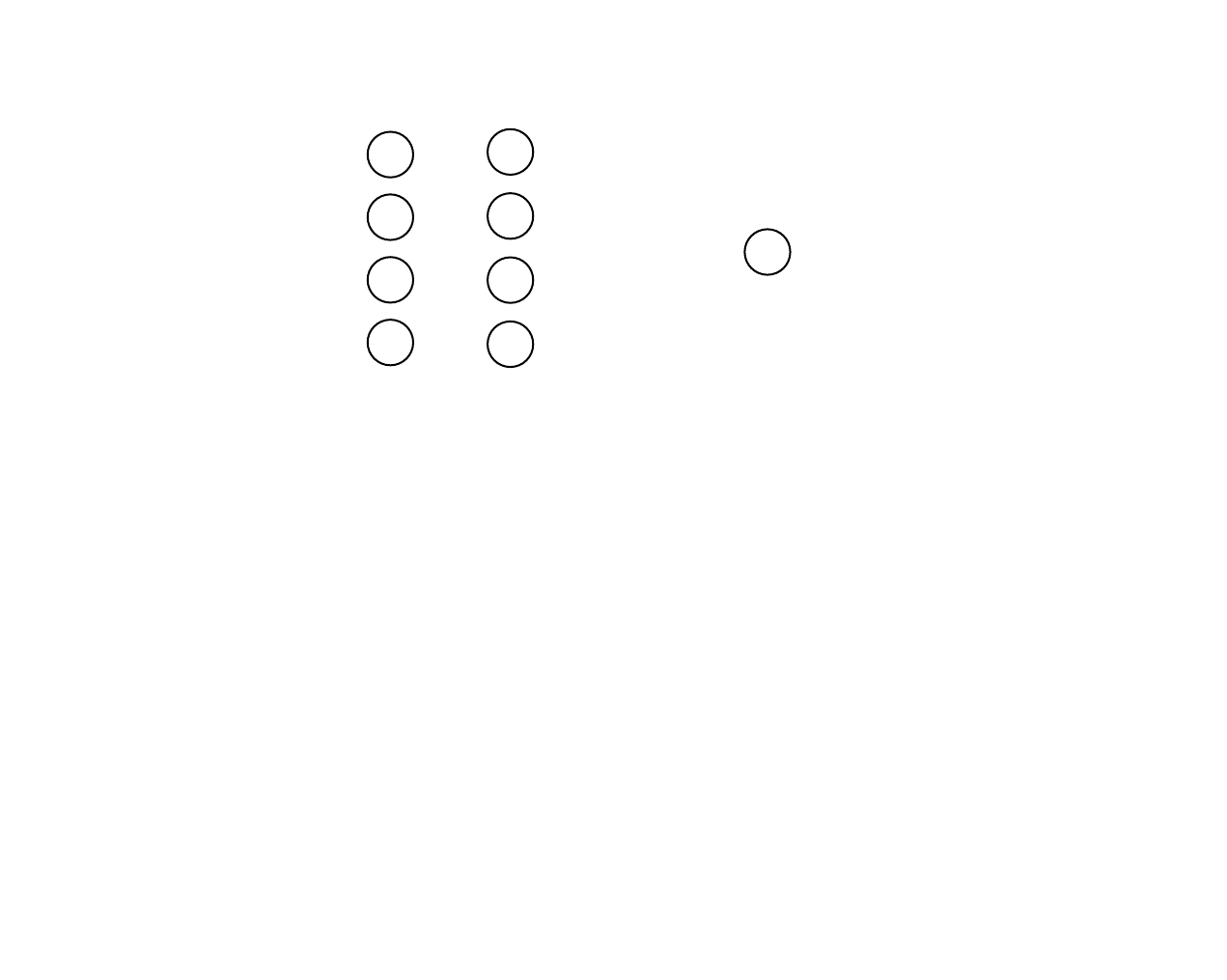}}%
    \put(0.03693518,0.51462415){\color[rgb]{0,0,0}\makebox(0,0)[lt]{\lineheight{1.25}\smash{\begin{tabular}[t]{l}State\end{tabular}}}}%
    \put(0,0){\includegraphics[width=\unitlength,page=2]{Overview11.pdf}}%
    \put(0.64172887,0.51007141){\color[rgb]{0,0,0}\makebox(0,0)[lt]{\lineheight{1.25}\smash{\begin{tabular}[t]{l}Action\\\end{tabular}}}}%
    \put(0,0){\includegraphics[width=\unitlength,page=3]{Overview11.pdf}}%
    \put(0.48309882,0.31704947){\color[rgb]{0,0,0}\makebox(0,0)[lt]{\lineheight{1.25}\smash{\begin{tabular}[t]{l}\textbf{Reward = ?}\end{tabular}}}}%
    \put(0.17527996,0.02352941){\color[rgb]{0,0,0}\makebox(0,0)[lt]{\lineheight{1.25}\smash{\begin{tabular}[t]{l}Environment\end{tabular}}}}%
    \put(0.5750587,0.72289974){\color[rgb]{0,0,0}\makebox(0,0)[lt]{\lineheight{1.25}\smash{\begin{tabular}[t]{l}RL Agent\end{tabular}}}}%
  \end{picture}%
\endgroup%

    \caption{\textbf{Reinforcement Learning Framework:} A RL agent uses a state vector to select an action that is implemented on a racing car. We ask the question of how to reward the agent based on the cars performance?}
    \label{fig:opti_obs_avoid}
\end{figure}

The paper starts by looking at related work in learning based methods for motion planning in Section \ref{sec:literature}.
In Section \ref{sec:theoretical} we present the theoretical design and discussion of reward signal design.
Section \ref{sec:prelims} reviews the navigation architecture stack that we develop and present our candidate solutions on in Section \ref{sec:sig_design}.
Sections \ref{sec:methodology} and \ref{sec:results} explain our evaluation methodology and present the results.

\section{Related Work} \label{sec:literature}

There have been many attempts to create certain kinds of robotic behavior using learning based method.
We start by studying mapless navigation methods \cite{tai2017virtual, chiang2018learning, zhu2017target, liu2017learning}, we briefly look at how imitation learning has been used \cite{zhang2016learning}, and then investigate current learning formulations for autonomous racing \cite{mnih2016asynchronous, de2018integrating, perot2017end, ivanov2020case}.

The problem of navigation is to move a mobile robot from a start position to and end goal without crashing.
Solutions to the navigation problem with RL have been well studied and it has been shown that agents can be trained to avoid obstacles from raw high-dimensional sensor readings, such as images \cite{zhu2017target}, and laser range finders \cite{tai2017virtual, chiang2018learning}.
Much of the the navigation literature is focused on the binary metric of reaching the target and thus rewards task completion and distance progress to the goal, while punishing crashing.
While such a reward signal is able suitable to generate task completion, it is not well suited to continuous metrics such as minimum time.

A solution to improving the long-horizon navigation performance of neural network planners is to use imitation learning (IL) to train a neural network to copy expert behavior.
A model predictive controller (MPC) with full state estimation, has been used as a teacher, to train a neural network to avoid obstacles \cite{zhang2016learning}.
While IL is able to train policies with good performance, it is limited to situations where an expert is available and thus undermines the aim of being able to learn from experience.

Several solutions have used RL agents as planners for autonomous racing.
The most common reward function which has been used to train autonomous agents for racing is some variation of rewarding velocity in the direction of the track and punishing distance from the center line \cite{mnih2016asynchronous, de2018integrating, perot2017end}.
This is a similar idea to the popular cross-track and heading controller from the control system literature.
Another reward which has been used is to punish the steering angle of the vehicle with the aim of creating smooth minimum curvature trajectories \cite{ivanov2020case}.

Reward signal design in navigation has been called a ``black art'' \cite{chiang2018learning}, and it is been recently noted that there remains a need to investigate how reward signals affect driving performance \cite{liu2017learning}.
We recognise the need to investigate the effect of reward signals on driving behaviors by studying the effect of different reward signals on autonomous racing performance.

\section{Theoretical Reward Design} \label{sec:theoretical}

Training a neural network is the process of transferring knowledge form one form to another, namely, an intended desired behavior to neural network weights and biases.
RL algorithms maximise the cumulative reward which the agent receives.
The intended desired behavior must therefore be well represented by the reward signal.

Different methods of rewarding RL agents are discussed in the context of applications where there is a minimum-time reference path that should be followed with certain additional un-mapped obstacles.
The two behavioral objectives that are considered are, the binary outcome of not crashing, and the continuous metric of time to complete the task.
To tackle these two objectives, a reward framework with terminal rewards for collision or completion are used in conjunction with an intermediate reward at each step to encourage high performance.
The aim is to design a reward signal that trains agents to produces high quality performance.

\begin{figure}[h]
    \centering
    \def \svgwidth{0.3\textwidth}
\begingroup%
  \makeatletter%
  \providecommand\color[2][]{%
    \errmessage{(Inkscape) Color is used for the text in Inkscape, but the package 'color.sty' is not loaded}%
    \renewcommand\color[2][]{}%
  }%
  \providecommand\transparent[1]{%
    \errmessage{(Inkscape) Transparency is used (non-zero) for the text in Inkscape, but the package 'transparent.sty' is not loaded}%
    \renewcommand\transparent[1]{}%
  }%
  \providecommand\rotatebox[2]{#2}%
  \newcommand*\fsize{\dimexpr\f@size pt\relax}%
  \newcommand*\lineheight[1]{\fontsize{\fsize}{#1\fsize}\selectfont}%
  \ifx\svgwidth\undefined%
    \setlength{\unitlength}{236.13072651bp}%
    \ifx\svgscale\undefined%
      \relax%
    \else%
      \setlength{\unitlength}{\unitlength * \real{\svgscale}}%
    \fi%
  \else%
    \setlength{\unitlength}{\svgwidth}%
  \fi%
  \global\let\svgwidth\undefined%
  \global\let\svgscale\undefined%
  \makeatother%
  \begin{picture}(1,0.5)%
    \lineheight{1}%
    \setlength\tabcolsep{0pt}%
    \put(0,0){\includegraphics[width=\unitlength,page=1]{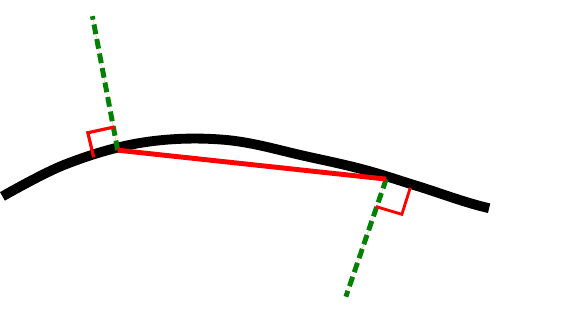}}%
    \put(0.3,0.18){\color[rgb]{0,0,0}\makebox(0,0)[lt]{\lineheight{1.25}\smash{\begin{tabular}[t]{l}$s_\text{t+1}-s_\text{t}$\end{tabular}}}}%
    \put(0,0){\includegraphics[width=\unitlength,page=2]{line_progress1.pdf}}%
  \end{picture}%
\endgroup%

    \caption{\textbf{Distance Based Reward:} The vehicles position is projected onto the line and used to measure the change in progress between time steps.}
    \label{fig:line_progress}
\end{figure}

The first methodology considered is to reward the agent based on the vehicles position at each timestep.
The position of the vehicle is used to reward the agent relative to the progress towards the goal.
The progress towards the goal can be measured in racing as the progress made along a line representing the race track.
Figure \ref{fig:line_progress} shows how the vehicles location is projected onto a line and then used to measure the progress.
The distance-based reward aims to maximise progress towards the target at each timestep.

\begin{figure}[h]
    \centering
    \def\svgwidth{0.3\textwidth}
\begingroup%
  \makeatletter%
  \providecommand\color[2][]{%
    \errmessage{(Inkscape) Color is used for the text in Inkscape, but the package 'color.sty' is not loaded}%
    \renewcommand\color[2][]{}%
  }%
  \providecommand\transparent[1]{%
    \errmessage{(Inkscape) Transparency is used (non-zero) for the text in Inkscape, but the package 'transparent.sty' is not loaded}%
    \renewcommand\transparent[1]{}%
  }%
  \providecommand\rotatebox[2]{#2}%
  \newcommand*\fsize{\dimexpr\f@size pt\relax}%
  \newcommand*\lineheight[1]{\fontsize{\fsize}{#1\fsize}\selectfont}%
  \ifx\svgwidth\undefined%
    \setlength{\unitlength}{263.84844467bp}%
    \ifx\svgscale\undefined%
      \relax%
    \else%
      \setlength{\unitlength}{\unitlength * \real{\svgscale}}%
    \fi%
  \else%
    \setlength{\unitlength}{\svgwidth}%
  \fi%
  \global\let\svgwidth\undefined%
  \global\let\svgscale\undefined%
  \makeatother%
  \begin{picture}(1,0.50772979)%
    \lineheight{1}%
    \setlength\tabcolsep{0pt}%
    \put(0,0){\includegraphics[width=\unitlength,page=1]{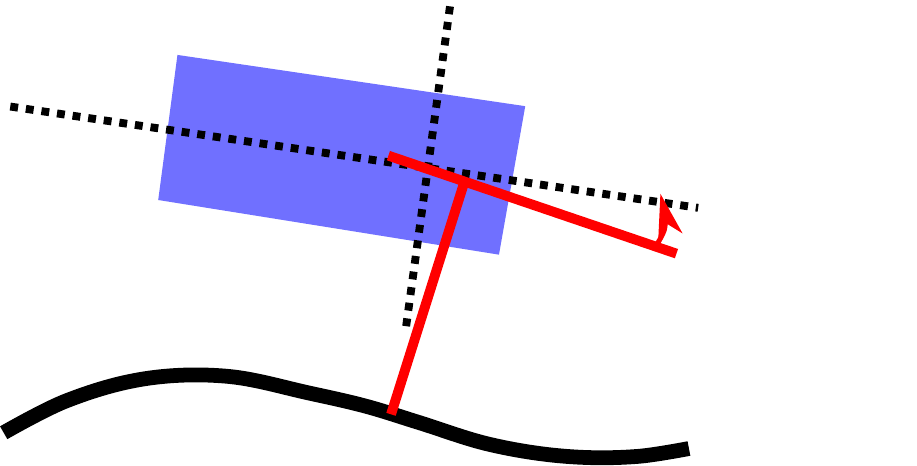}}%
    \put(0.74698444,0.22920404){\color[rgb]{0,0,0}\makebox(0,0)[lt]{\lineheight{1.25}\smash{\begin{tabular}[t]{l}$\theta$\end{tabular}}}}%
    \put(0.46597807,0.11700662){\color[rgb]{0,0,0}\makebox(0,0)[lt]{\lineheight{1.25}\smash{\begin{tabular}[t]{l}$d_\text{c}$\end{tabular}}}}%
    \put(0,0){\includegraphics[width=\unitlength,page=2]{cth21.pdf}}%
  \end{picture}%
\endgroup%

    \caption{\textbf{Cross-track \& Heading Reward:} Illustration of how cross-track distance, $d_\text{c}$, and heading error, $\theta$, are measured}
    \label{fig:cth_controller}
\end{figure}

The derivative of position, velocity, provides valuable information about where the vehicles position in the next timestep.
Therefore, a reward based on the position and velocity is developed with an aim to better improve the long-term performance of the vehicle.
Taking the velocity into account ensures that the planner is concerned with where the vehicle is moving in the future, in addition to where it currently is.
A reward which has previously been used in racing is to reward the vehicle for its velocity along a reference line and punish the vehicles lateral deviation from the line.
The velocity along the line is calculated according to the cosine of the angle difference between the vehicle and the reference, and the distance is simply measured.
Figure \ref{fig:cth_controller} shows how the cross-track distance, $d_c$, and heading error, $\theta$, are measured.
This reward signal uses the velocity of the vehicle in addition to the vehicle's position to calculate the reward.

The third methodology presented is to reward the agent based on the actions that are selected.
In racing, minimum curvature trajectories are usually similar to time optimal trajectories since one of the dominant constraints is the lateral friction limit of the tyres and the track.
The minimum curvature path around a race track is the one in which the vehicle steers as little as possible.
We therefore consider a reward signal that punishes steering with the aim of generating minimum curvature paths.


In the existing literature, the distance based and cross-track \& heading error reward equations always use the center line of the race track as the reference line.
We introduce the novel idea of using the minimum curvature path as a reference for the reward since minimum curvature paths generally produce faster lap times than merely following the center line.
It is thus expected that using the minimum curvature path as the reference for the reward signal will improve the vehicles time performance.


\section{Preliminaries: Navigation Architecture} \label{sec:prelims}

The reward signals are evaluated in using a standard perception, planning and control stack \cite{perceptionPlanningControl}.
A global planner is used to generate a minimum curvature trajectory around a track with a minimum time speed profile \cite{heilmeier2019minimum, lipp2014minimum}.
The planner uses the global plan and current sensor readings to determine speed and steering references.
A low level proportional control system executes the navigation references on the hardware.

\begin{figure}[h]
    \centering
    \usetikzlibrary{shapes.geometric, arrows}

\usetikzlibrary{shapes, shadows, arrows, arrows.meta}

\tikzstyle{empty} = [rectangle, rounded corners, draw=white, minimum height = 0.2cm, minimum width=0.2cm]
\tikzstyle{block} = [rectangle, rounded corners, draw=black, minimum height = 1cm, minimum width=2cm]
\tikzstyle{mysum} = [circle, draw=black]

\tikzstyle{arrow} = [thick, ->, >=stealth]


\begin{tikzpicture}[node distance=2cm]

    \node (base) [empty] {};
    \node (b2) [empty, below of=base] {};
    \node (pf) [block, right of=base, xshift=0.4cm] {Path Follower};
    \node (nn) [block, below of=pf, xshift=2.5cm] {Neural Network};
    \node (sum) [mysum, right of=pf, xshift=2.5cm] {};
    \node (b3) [empty, right of=sum, xshift=-18] {};

    \draw [arrow] (pf) -| (nn);
    \draw [arrow] (nn) -| (sum);
    \draw [arrow] (pf) -- (sum);
    \draw [arrow] (base) -- (pf);
    \draw [arrow] (b2) -- (nn);
    \draw [arrow] (b2) -| (pf);
    \draw [arrow] (sum) -- (b3);
    
    \node[align=left] at (0.4, 0.6) {Global \\ Plan};
    \node[align=left] at (0.4, -1.5) {Vehicle \\ state};
    \node at (6.4, 0.2) {$+$};
    \node at (6.7, -0.5) {$+$};
    \node[align=left] at (7.8, -0.6) {Modified \\ references};

\end{tikzpicture}

    \caption{Modification planner showing how a path follower and neural network are used in parallel to avoid obstacles while maintaining a reference trajectory.}
    \label{fig:mod_arch}
\end{figure}
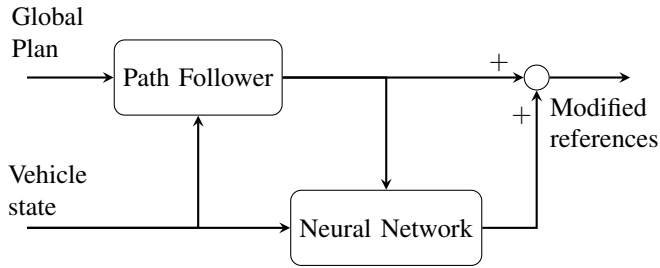

Figure \ref{fig:mod_arch} shows the \textit{modification planner} presented in \cite{evans2021autonomous} that is used.
The planner uses a path follower, in parallel with a neural network to follow a reference trajectory while avoiding obstacles. 
The path follower follows a precalculated reference trajectory using the pure pursuit method with a fixed look-ahead distance \cite{coulter1992implementation}.
The neural network receives the path follower references and the state of the vehicle and is able to modify the steering reference to avoid obstacles.

The neural network is trained with reinforcement learning to maximise a reward signal.
At each step, the network receives a state vector comprising of, the current velocity and steering commands, the path followers calculated velocity and steering commands, and the readings from the laser range finders.
The neural network (nn) then selects an action which is used to modify the path follower (pf) references according to, $\delta_\text{ref} = \delta_\text{pf} + \delta_\text{nn}$
After each action has been taken, a reward is calculated for the state-action pair.

\section{Reward Signal Equation Design} \label{sec:sig_design}

The ideas which were presented in Section III are now developed into equations that can easily be calculated at each step during training.
We express the framework for all our reward signals which punishes the agent for crashing and rewards the agent for completing a lap as,
\begin{equation}
    r(\textbf{s}_\text{t}, \textbf{a}_\text{t}) = 
    \begin{cases}
        r_{\text{crash}} = -1 & \quad \quad \text{if crash} \\
        r_\text{complete} = 1 & \quad \quad \text{if lap complete} \\
        r_\text{racing}  & \quad \quad \text{otherwise}, \\
    \end{cases}
\end{equation}
where $r_\text{racing}$ is the reward that we propose to encourage optimal trajectories.
All of our rewards are scaled to be independent of vehicle or track used.

\subsection{Baseline Reward: No Racing Reward}

The baseline reward is to give no additional reward to influence the trajectory, $r_\text{racing} = 0 $.
Under this system, the agent will simply learn not to crash and to finish laps.

\subsection{Distance-Based Reward}

The common navigation reward based on distance to a goal is amended to be used for racing on a track to use the progress the vehicle has made along the race track.
We scale the progress made between each time step according to the total track length.
We write the \textit{Distance reward} as,
\begin{equation}
    r_\text{racing} = \beta_\text{distance} \frac{(s_\text{t+1} - s_\text{t})}{s_\text{total}}
\end{equation}
where $s_\text{t}$ is the progress along the track at time $t$.
The hyper-parameter $\beta_\text{distance}$ is the total amount of additional reward which the agent will receive for completing the track.
We use both the track center line and minimum curvature path as a reference line.

\subsection{Cross-track \& Heading (CTH) Reward}

Racing behavior has been generated by using a reward signal that rewards velocity along the track center line and punishes the lateral deviation from the center line.
The cross-track \& heading reward is written as,
\begin{equation}
    r_\text{racing} = \beta_\text{heading} ~ V_\text{t} \cos \theta - \beta_\text{cross-track} ~ d_\text{c},
\end{equation}
where, $V_\text{t}$ is the speed of the vehicle, $\theta$ is the heading error and $d_\text{c}$ is the cross-track error.
The velocity is scaled according to the vehicles maximum velocity and the cross-track distance according to the width of the track.
Once again, the track center line and minimum curvature paths are used.

\subsection{Minimum Steering Reward}

We attempt to generate minimum curvature paths around obstacles by punishing the magnitude of the total steering action. 
It is aimed for and expected that the RL algorithm will learn a policy which generates trajectories with the lowest curvature.
The minimum steering reward signal is written as, 
\begin{equation}
    r_\text{racing} = - \beta_\text{steering}~ |\delta_\text{ref}|,
\end{equation}
where $\delta_\text{ref}$ is the steering angle which the local planner outputs.
The steering angle is scaled according to the vehicles maximum steering angle.

\section{Experimental Methodology} \label{sec:methodology}

\subsection{Simulation Environment}

We train and evaluate our vehicles in the context of F1/10th autonomous racing.
A simulator was custom built so that the state of the vehicle on the map could be easily accessed and used in the reward signals.
\footnote{Our simulation code is available online at: \url{https://github.com/BDEvan5/RewardSignalDesign}}
The simulator is modelled on the OpenAI-Gym environment to take an action at each time step, update the state according to stationary transition dynamics, and then return the new state.

An F1/10th car is simulated using the kinematic bicycle model \cite{kinematic_bicycle_model} as in common in similar simulators \cite{o2020textscf1tenth}.
The simulator takes an action in the form of a velocity and steering command. 
A proportional control system implements the velocity and steering commands by calculating and executing the acceleration and the change in steering required.
Table \ref{tab:sim_params} lists several important simulation parameters.

\begin{figure}[h]
    \centering
    \def\svgwidth{0.28\textwidth}
    \input{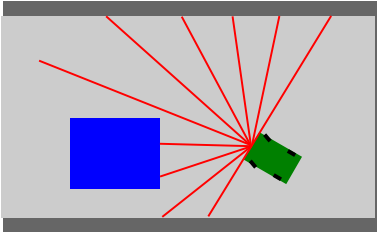}
    \caption{Example scenario from simulator showing a non-holonomic vehicle on a track and the relevant range finder readings. The range finders are equally spaced in front of the vehicle and limited at a maximum range}
    \label{fig:simulator}
\end{figure}

The simulator returns a state consisting of the location, bearing, steering and velocity of the vehicle on the map and 10 sparse laser range finder readings.
Figure \ref{fig:simulator} shows an image of the vehicle and range finders on a section of race track.

We train and test our agents on a standard F1/10th map, Porto, since it has a simple form.
The track, shown in figure \ref{fig:cth_ref_noObs}, is received as an occupancy grid and then converted to a set of center line points with corresponding track widths and normal vectors.
The center line points and track widths are used for the global trajectory optimisation, which produces a set of time optimal way points \cite{heilmeier2019minimum}.

\begin{table}[h]
    \centering
    \begin{tabular}{c|c}
        \textbf{Simulation Parameter} & \textbf{Value} \\
        \hline
        Number of range finders & 10 \\
        Max range finder value & 4 m\\
        Obstacle size & 0.6 m \\
        Simulation time step & 0.01 s \\
    \end{tabular}
    \caption{Parameters used in the simulation}
    \label{tab:sim_params}
\end{table}

Each episode consists of a single lap of the track or until the agent crashes.
At the beginning of each episode, three or four obstacles are randomly spawned along the track. 
We use square obstacles with a length of 0.6m which is a similar size to an F1/10th car.

\subsection{Comparative Baseline Solutions}

We compare our solutions to current successful planning strategies. 
For Benchmark 1, we compare our work to a standard pure pursuit path follow that follows the precalculated optimal trajectory.
The reference trajectory given to the path follower is the same as used in the network-based planners, and uses a fixed look-ahead distance to calculate the steering angle.

For Benchmark 2, we compare our solutions to the popular ``Follow the Gap Method'' (FGM) \cite{sezer2012novel}.
The FGM identifies a bubble around the nearest obstacle and then navigates away from it into free space.

\subsection{Neural Network Training} 

We select a training regime that is used to train all of the local planners.
The training regime is a controlled variable that is kept constant for all the local planners that are evaluated.

The Twin Delayed Deep Deterministic Policy Gradient algorithm (TD3), is used as the reinforcement learning algorithm \cite{td3_paper}.
The TD3 algorithm was selected because it is currently one of the best RL algorithms for continuous control.
The original hyper-parameters for the algorithm are used in our implementation.

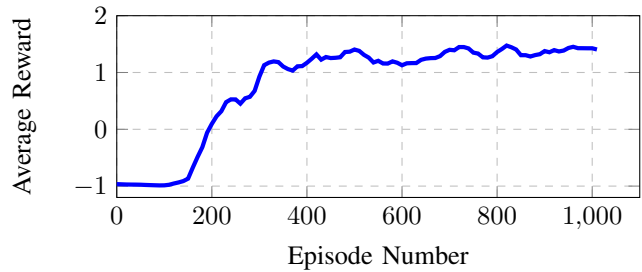
\begin{figure}[h]
    \centering


\begin{tikzpicture}
	\begin{axis}
		[name=plot,width=.48\textwidth,
		height=4cm,
		xmin=0, xmax=1100,
		ymin=-1.2, ymax=2,
		grid=major,
		grid style=dashed,
		xlabel={Episode Number},
		ylabel={Average Reward}]

		\pgfplotstableread[col sep=comma,]{Graphs/TrainingData.csv}{\table}
		
		\addplot[color=blue, ultra thick] table[x={x}, y={y}]{\table};
		
		\label{training_plot}
	
	\end{axis}
\end{tikzpicture}

    \caption{Example training graph using the distance reward}
    \label{fig:train_data}
\end{figure}

We use simple neural networks that consist of two fully connected hidden layers of 200 neurons each.
After each hidden layer, the \textit{ReLu} activation function is applied.
The final action is passed through the \textit{tanh} activation function to give an output in the range [-1, 1].
The neural networks are trained in mini-batches of 100 samples for a total of 100,000 steps.
This number was selected as it was shown that the networks have all converged by this many steps.
Figure \ref{fig:train_data} shows a typical training graph for one of the agents trained.

The hyper-parameter values for each reward signal were manually tuned to achieve high completion rates.
Table \ref{tab:hps} notes the hyper-parameter values that we use so that our results may be reproduced.

\begin{table}[h]
    \centering
    \begin{tabular}{c|c}
        \textbf{Hyper-parameter} & \textbf{Value} \\
        \hline
        $\beta_\text{distance}$ & 0.5 \\
        $\beta_\text{heading}$ & 0.04 \\
        $\beta_\text{cross-track}$ & 0.004 \\
        $\beta_\text{steering}$ & 0.01 \\
        
    \end{tabular}
    \caption{Hyper-parameter values selected for reward signals}
    \label{tab:hps}
\end{table}

\section{Results} \label{sec:results}

\subsection{Benchmark 1: Performance Without Obstacles}

Testing the agents on tracks with no obstacles shows how well the planner is able to retain its ability to follow a path when no obstacles are present.
Table \ref{tab:results_no_obs} shows the results from each of the agents being tested on a track with no obstacles.

\begin{table}[h]
    \centering
    \begin{tabular}{p{3cm}|c}

        \textbf{Vehicle/Reward Name} & \textbf{Time (s) }  \\
        \hline
        \hline
        Reference Path Follower  &  7.2    \\
        No Racing Reward & 10.7 \\
        \hline
        Distance (center line) & 8.5 \\
        Distance (min. curve line)  & 9.6 \\
        CTH (center line) & 9.5 \\
        CTH (min. curve line)  & 8.5 \\
        Minimum steering &  9.8 \\
    \end{tabular}
    \caption{Results from agents tested on the Porto race track with no obstacles}
    \label{tab:results_no_obs}
\end{table}

All of the racing reward signals improved on the baseline of no racing reward by producing faster lap times.
The racing reward signals also lead the vehicle to crash marginally more often.
The pure pursuit solutions is presented as a baseline of the best time possible for the neural network-based local planners.

\begin{figure}[h]
    \centering
    \def\svgwidth{0.4\textwidth}
    \input{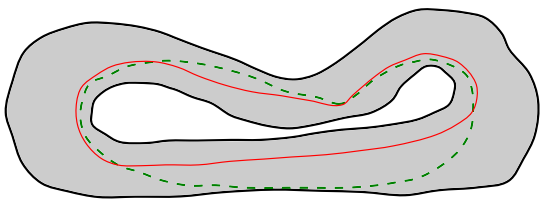}
    \caption{Example trajectory of a vehicle trained with the CTH (min. curve line) reward signal and tested on a track with no obstacles. The grey area is the track, the green dashed line is the reference and the red line is the path taken. The vehicle closely follows the reference in the corners and deviates more during the straight section.}
    \label{fig:cth_ref_noObs}
\end{figure}

The \textit{distance (center line)} and \textit{CTH (min. curve line)} rewards performed the best by achieving the lowest average lap time of 8.5 seconds.
Figure \ref{fig:cth_ref_noObs} shows a typical trajectory taken by the \textit{CTH (min. curve)} reward signal.
The vehicle learns to follow the reference trajectory closely, especially around the corners of the track.

\begin{figure}[h]
    \centering
    \def\svgwidth{0.4\textwidth}
    \input{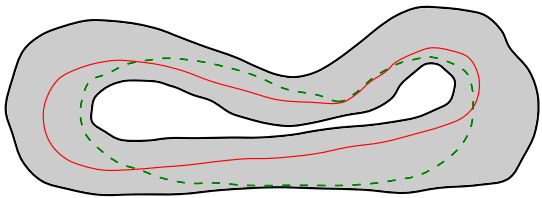}
    \caption{Example trajectory of vehicle trained with the steering reward racing in Benchmark 1. The trajectory shows how the vehicle stays in the middle of the track and takes a long path through the corners.}
    \label{fig:steer}
\end{figure}

The \textit{distance (min. curve line) reward} and the \textit{CTH (center line) reward} achieved times of 9.5 and 9.6 seconds respectively. 
The \textit{minimum steering reward} was the slowest of the racing rewards with a time of 9.8 seconds.
Figure \ref{fig:steer} shows how the behavior which is learned by the \textit{minimum steering reward} appears to minimise the steering at each step which results in the corners being taken very poorly.

A trend which was observed to varying degrees in all neural network planners is that they avoid the edges of the track, which causes them to take turns more sharply and thus more slowly.
All of the example trajectories show how the path taken by the vehicle is further away from the edge than the reference trajectory.

\subsection{Benchmark 2: Performance with Obstacles}

Table \ref{tab:results_wObs} shows the results of the agents on the race track that they were trained on tracks with randomly spawning obstacles.
We tested our agents by running 1000 laps with each of them and present the average times and \% completion rate.

\begin{table}[h]
    \centering
    \begin{tabular}{p{3cm}|c c }

        \textbf{Vehicle} & \textbf{Avg. Time} & \textbf{\% Complete}  \\
        \hline
        \hline
        Follow The Gap & 9.8   & 84.0\% \\
        No Racing Reward &  11.1   & 99.8\% \\
        \hline
        Distance (center line) & 10.0    & 95.8 \\
        Distance (min. curve line) & 9.9   & 98.3 \\        
        CTH (center line) & 9.4   & 99.9 \\
        CTH (min. curve line)  & 8.8   &  94.3\% \\
        Minimum steering & 10.2  &  97.0\%    \\
        
    \end{tabular}
    \caption{Results from agents tested on race track with random obstacles}
    \label{tab:results_wObs}
\end{table}

The Benchmark 2 results, show that all of our proposed reward signals are able to decrease the lap time than not using a racing reward.
Our solutions are able to compete with the Follow the Gap Method with two of the agents achieving faster average times.

\begin{figure}[h]
    \centering
    \def\svgwidth{0.4\textwidth}
    \input{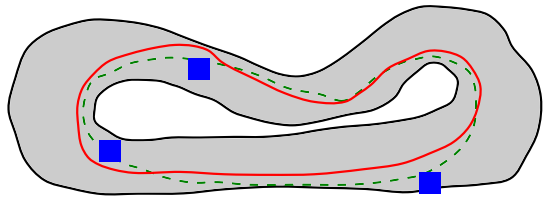}
    \caption{Example Trajectory of CTH (min. curve line) avoiding obstacles.}
    \label{fig:cth_glbl_avoid}
\end{figure}

The \textit{CTH (min. curve line)} reward signal achieves the best average time of 8.8 seconds and lowest completion rate of 94.3\%.
Figure \ref{fig:cth_glbl_avoid} shows how the \textit{CTH (min. curve line)} vehicle typically maintains the reference path, while deviating to avoid obstacles.

The \textit{CTH (center line) reward} comes a close second with an average time of 9.4 seconds.
The \textit{distance (center line) reward} produces and average time of 10.0 and the \textit{distance (min. curve line) reward} produces an average time of 9.9. 
It is also noted that both of these reward signals have good completion rates of 99.9 and 98.3\%.
These results show that using the progress along a line as a reward produces improved racing behavior while maintaining a high level of safety.

\begin{figure}[h]
    \centering
    \def\svgwidth{0.4\textwidth}
    \input{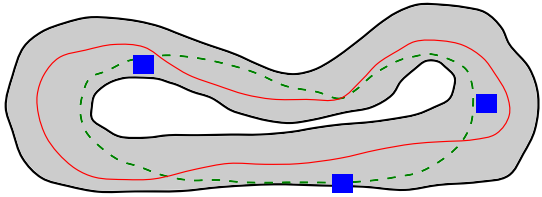}
    \caption{Example trajectory from the minimum steering reward shows the agent takes a long path to avoid obstacles.}
    \label{fig:strVel}
\end{figure}

The \textit{minimum steering reward} achieved the slowest average time of the racing rewards test of 10.2 seconds.
Figure \ref{fig:strVel} shows the typical behavior from the \textit{minimum steering reward} and how it takes a long path around obstacles which results in slower lap times.

\subsection{Discussion}

The cross-track and heading (min. curve line) reward signal has shown to outperform the other reward candidates by achieving the fastest lap times with and without obstacles.
Using a minimum curvature line as a reference was shown to produce faster lap times than the center line as the reward signal trains the network to closely follow the reference line while avoiding obstacles.

Using the distance along the reference line is effective and able to produce good racing behavior.
It is suggested that the reason that it is slower than the cross-track error signal is because it only encodes the position of the vehicle in the reward as opposed to position and velocity.
While the minimum steering reward signal is the worst performer of the rewards considered, it still improves the racing performance.


The work presented here should be extended to designing neural network-based agents that are capable of competitive head-to-head racing.
It is expected that this will involve stacking multiple states and a reward signal that takes being in the lead into account.
A long standing question that should be further investigated is the safe application of neural network based planners.
Safety based reward signals should be explored to see how the safety of neural network based planners can be improved.

\section{Conclusion}

In this paper, the question of reward signal design for robotic systems with multiple, continuous, competing metrics was studied.
The three different reward methodologies of rewards based on the position of the vehicle, direction of the velocity and action selected were proposed and evaluated in the context of F1/10th autonomous racing.
The reward signal candidates were evaluated in the context of F1/10th autonomous racing and the results showed that the cross-track \& heading reward signal generated the fastest average lap times while using no racing reward resulted in the fewest crashes.
The center line and a minimum curvature line were used as the references for the position and velocity based reward signals and the results indicated that uses a line of a minimum curvature leads to faster lap times.
These results contribute to better understanding how reward signals can be designed to generate specific kinds of behavior in robotic systems.

\typeout{}


\end{document}